\documentclass[11pt,a4paper]{article}
\usepackage{times}
\usepackage{latexsym}
\usepackage[utf8]{inputenc}
\usepackage{todonotes}
\usepackage{booktabs}
\usepackage{multirow}
\usepackage{arydshln}
\usepackage{adjustbox}
\usepackage{tikz}
\usepackage{gnuplot-lua-tikz}
\usepackage{microtype}
\usepackage{stfloats}

\usepackage{amsmath}
\usepackage{amssymb}
\usepackage{algorithm}
\usepackage[noend]{algpseudocode}
\usepackage{float}
\newfloat{algorithm}{t}{lop}

\PassOptionsToPackage{breaklinks}{hyperref}
\usepackage{xurl}
\usepackage[hyperref]{acl2019}

\def\Pr{\ensuremath\mathsf{P}}

\def\O{{\hphantom 0}}

\aclfinalcopy 
\def\aclpaperid{3196} 

\usepackage[normalem]{ulem}

\def\ODdel#1{\bgroup\markoverwith{\textcolor{cyan!89!yellow!80!black!100}{\rule[0.4ex]{2pt}{3pt}}}\ULon{#1}}

\title{Improving Fluency of Non-Autoregressive Machine Translation}

\author{Zdeněk Kasner \and Jindřich Libovický \and Jindřich Helcl \\
  Charles University, Faculty of Mathematics and Physics, \\
  Institute of Formal and Applied Linguistics, \\
  Malostranské náměstí 25, 118 00 Prague, Czech Republic \\
  \texttt{\{kasner, libovicky, helcl\}@ufal.mff.cuni.cz}
}

\date{}

\begin{document} \maketitle \begin{abstract}
%
%
Non-autoregressive (nAR) models for machine translation (MT) manifest superior
decoding speed when compared to autoregressive (AR) models, at the expense of
impaired fluency of their outputs.
We improve the fluency of a nAR model with connectionist temporal
classification (CTC) by employing additional features in the scoring model used
during beam search decoding.
Since the beam search decoding in our model only requires to run the network in
a single forward pass, the decoding speed is still notably higher than in
standard AR models.
We train models for three language pairs: German, Czech, and Romanian from and
into English.
The results show that our proposed models can be more efficient in terms of decoding speed and still achieve a competitive BLEU score relative to AR models.
\end{abstract}

\section{Introduction}


One of the challenges that the research community faces today is improving the
latency of neural machine translation (NMT) models. The decoders in modern NMT
models operate autoregressively, which means that the target sentence is
generated in steps from left to right
\citep{bahdanau2015neural,vaswani2017attention}. In each step, a token is
generated and it is supplied as the input for the next step.



Recently, nAR models for NMT tackled this issue by reformulating
translation as sequence labeling. As long as the model and the data
fit in a GPU memory, all computation steps can be done in parallel
\citep{gu2017nonautoregressive, lee2018deterministic, libovicky2018end,
ghazvininejad-etal-2019-mask}. However, such models suffer from less fluent outputs.


In phrase-based statistical machine translation (SMT\@;
\citealp{koehn2009statistical}), the translation fluency is handled by a
language model component, which is responsible for arranging the phrases
selected by the decoder into a coherent sentence. In AR NMT\@, there is no
external language model. The decoder part of the neural model plays the role of
a conditional language model, which estimates the probability of the
translation given the source sentence signal as processed by the encoder part.


In automatic speech recognition (ASR), \citet{graves2014towards} proposed a
beam search algorithm which combines an $n$-gram language model with scores
from a model trained using CTC (\citealp{graves2006connectionist}). 

In this paper, we adopt and generalize this approach for nAR NMT by extending
a CTC-based model by \citet{libovicky2018end}. 
We experiment with these models on six language pairs and we find
that the generalized decoding algorithm helps narrowing the performance gap between the
CTC-based and the standard AR models.


\section{Non-autoregressive MT with CTC}%
\label{sec:ctc}

Non-autoregressive models for MT formulate the translation problem as sequence
labeling. The states of the final decoder layer are independently labeled with
target sentence tokens.
The models can parallelize all steps of the computation and thus reduce the decoding time substantially. The nAR
models were enabled by the invention of the self-attentive Transformer model
\citep{vaswani2017attention}, which allows arbitrary reordering of the states
in each layer. Most of the nAR models need a prior estimate of
the sentence length, either explicitly \citep{lee2018deterministic} or via a
specialized fertility model \citep{gu2017nonautoregressive} and rely on the
attention mechanism for re-ordering.

We base our work on an alternative approach that does not depend on the target
length estimation. Instead, it constrains the upper bound of the target
sentence length to the source sentence length multiplied by a fixed number $k$
and uses CTC to compute the training loss \citep{libovicky2018end}.

The architecture consists of three components: encoder, state splitter, and
decoder. The encoder is the same as in the Transformer model. The state
splitter takes each state from the final encoder layer and projects it into $k$
states of the original dimension, making the sequence $k$ times longer. The
decoder consists of additional Transformer layers which attend to both
encoder and state splitter outputs.
%
%





CTC enables the model to generate variable-length sequences using a special
\emph{blank symbol} that is included in the vocabulary. The resulting training
loss is a sum of cross entropy of all possible interleavings of the reference
sequence with the blank symbols. Even though enumerating all the combinations
is intractable, the cross-entropy sum can be efficiently computed using a
dynamic programming forward-backward algorithm.

Since each token can be decoded independently of other tokens at inference
time, the model reaches a significant speedup over the AR models. However, this
speedup is achieved at the expense of the translation quality which manifests
mostly in the reduced fluency.

\section{Proposed Method}%
\label{sec:method}

We tackle the reduced fluency problem using beam search and employing
additional features in its scoring model. Our approach is inspired by
statistical MT and ASR\@.

\subsection{Beam Search with CTC}

\begin{algorithm}
\caption{Beam Search Algorithm with CTC}\label{alg:beam}
\begin{algorithmic}[1]

\State $\mathcal{B} \gets \{\varnothing\}$ \Comment{ {\small Beam} }

\For{step $i$ = $1 \ldots k \cdot T_x$} 

\State $H \gets \varnothing$  \Comment{ {\small Hypothesis $\rightarrow$ CTC score} }
\State $W \gets 2n\text{-best tokens in step }i$

\For{hypothesis $h \in \mathcal{B}$} 
\For{token $w \in W$} 

\State $s \gets \Pr_i(w) \cdot \Pr(h)$ \label{alg:beam:prod} \Comment{ {\small derivation score }}
\State $H[h+w] \gets H[h+w] + s $ \label{alg:beam:sum}

\EndFor
\EndFor

\State $\mathcal{B} \gets \textbf{select\_nbest}(H, n)$ \label{alg:beam:nbest}

\EndFor

\State \Return$\mathcal{B}$

\end{algorithmic}
\end{algorithm}

Unlike greedy decoding, which can be performed in parallel by selecting
tokens with the highest probability in each step independently, beam search operates sequentially.
However, the speedup gained from the parallelization is preserved because the output
probability distributions are still conditionally independent and thus can be 
computed in a single pass through the network -- as opposed to the AR models, 
which need to re-run the entire stack of decoder layers every step.

The beam search algorithm for the CTC-based model \citep{graves2014towards} is shown
in Algorithm~\ref{alg:beam}. Unlike standard beam search in NMT,
the algorithm needs to deal with the issue that a single hypothesis may have various
derivations, depending on the positions of the blank symbols. The score of a
single derivation is the product of the conditionally independent probabilities
of the output tokens (line~\ref{alg:beam:prod}).
%

The beam search score of a hypothesis is then the sum of the scores of its
derivations formed in the current beam search step (line~\ref{alg:beam:sum}).

\subsection{Scoring Model}

For selecting $n$ best hypotheses (line~\ref{alg:beam:nbest} in
Algorithm~\ref{alg:beam}), we employ a linear model to compute the score:
\begin{equation}
	\textit{score} = \log\Pr(y | x) + \mathbf{w} \cdot \mathbf{\Phi}(y)
\end{equation}
where $\Pr(y | x)$ is the CTC score of the generated sentence $y$ given a
source sentence $x$, $\mathbf{\Phi}$ is a feature function of $y$ and
$\mathbf{w}$ is a trainable feature weight vector.

We use structured perceptron for beam search to learn the feature weights
\citep{huang2012structured}. During training, we run the beam search algorithm
and if the reference translation falls off the beam, we apply the perceptron
update rule:
\begin{equation}
	\mathbf{w} \leftarrow \mathbf{w} + \alpha \left( \mathbf{\Phi}(y) -
	\mathbf{\Phi}(\hat{y}) \right)
\end{equation}
where $\alpha$ is the learning rate, $\mathbf{\Phi}(y)$ are the feature values
of the prefix of the reference translation in the given time step, and
$\mathbf{\Phi}(\hat{y})$ are the feature values of the highest-scoring
hypothesis in the beam. Alternatively, we found that applying the perceptron
update rule multiple times with all hypotheses that scored higher than the
reference leads to faster convergence.
In order to stabilize the training, we
do not train the weight of the CTC score and set it to 1.

In the following paragraphs, we describe the features $\mathbf{\Phi}$
used within our beam search algorithm.

\paragraph{Language Model.} 
The main component improving the fluency is a language model (LM). For
efficiency, we use an $n$-gram LM\@. Since the hypotheses contain blank
symbols, the beam may consist of hypotheses of different lengths. Because
shorter sequences are favored by the LM, we divide the log-probability of each
hypothesis by its length in order to normalize the scores.

\paragraph{Blank/non-blank symbols.} 
To guide the decoding towards sentences of correct length, we compute the ratio
of blank vs.\ non-blank symbols as follows:
\begin{equation*}
	\max\left(0,  \frac{\text{\# blanks}}{\text{\# non-blanks}} - \delta
	\right)
\end{equation*}
where $\delta$ is a hyperparameter that thresholds the penalization for too high
blank/non-blank symbol ratio. Based on the distribution properties of the
ratio, we use $\delta = 4$.

\paragraph{Trailing blank symbols.} 
We observed that the outputs produced by the CTC-based model tend to be too short. To
prevent that, we count the trailing blank symbols:
\begin{equation*}
   \max\left(0,  \text{\# trailing blanks} - \text{source length} \right).
\end{equation*}

\section{Experiments}%
\label{sec:experiments}

\begin{table*}[ht]

    \begin{center}
    \begin{adjustbox}{max width=0.95\textwidth}
    \begin{tabular}{lccccccc}
    \toprule
    \multirow{2}{*}[-2pt]{Method} &
    \multicolumn{2}{c}{German WMT15} &
    \multicolumn{2}{c}{Romanian WMT16} &
    \multicolumn{2}{c}{Czech WMT18} &
    \multirow{2}{2cm}[-2pt]{\centering Decoding time [ms]}
    \\ \cmidrule(lr){2-3} \cmidrule(lr){4-5} \cmidrule(lr){6-7}

    & en $\rightarrow$ de
    & de $\rightarrow$ en
    & en $\rightarrow$ ro
    & ro $\rightarrow$ en
    & en $\rightarrow$ cs
    & cs $\rightarrow$ en

    \\ \midrule

%
%
%
%
%

    Non-autoregressive
      & 21.67 
      & 25.57   
      & 19.88   
      & 28.99   
      & 16.27   
      & 17.63   
      & \O233
    \\
    Transformer, greedy
      & 29.84 
      & 32.62 
      & 25.89 
      & 33.54 
      & 21.57 
      & 27.89 
      & 1664
    \\
    Transformer, beam 5
      & 30.23 
      & 33.43 
      & 26.46 
      & 34.06 
      & 22.20 
      & 28.49 
      & 3848
    \\
    \midrule

    Ours, beam 1
    & 22.68  
    & 26.44  
    & 19.74  
    & 29.65  
    & 16.98  
    & 18.78  
    & \O337

      \\
    Ours, beam 5
    & 25.50  
    & 29.45  
    & 22.46  
    & 33.01  
    & 19.31  
    & 23.33  
    & \O408
      \\
    Ours, beam 10
    & 25.93  
    & 30.05  
    & 23.33  
    & 33.29  
    & 19.47  
    & 23.95  
    & \O526

      \\
    Ours, beam 20
    & 26.03  
    & 30.15  
    & 24.11  
    & 33.51  
    & 19.58  
    & 24.32  
    & 1097

      \\
    \bottomrule
    \end{tabular}
    \end{adjustbox}
    \end{center}

    \caption{Quantitative results of the models in terms of BLEU score and
    average decoding times per sentence in milliseconds. Results on WMT14 English-German translation and results without back-translation are in the Appendix.}%
    \label{tab:results}

\end{table*}

We perform experiments on three language pairs in both directions:
English-Romanian, English-German, and English-Czech.

For training the base NMT models, we use WMT parallel
data,\footnote{http://statmt.org/wmt19/translation-task.html} which
consists of 0.6M sentences for English-Romanian, 4.5M sentences for
English-German, and 57M sentences for English-Czech.

Further, we use the WMT monolingual data: 20M sentences for English, German and
Czech and 2.2M sentences for Romanian for training the LM and for
back-translation.

We preprocess all data using
SentencePiece\footnote{https://github.com/google/sentencepiece}
\citep{kudo2018sentencepiece}. We train the SentencePiece models with a
vocabulary size of 50,000.

We implement the proposed architecture using Neural
Monkey\footnote{https://github.com/ufal/neuralmonkey} \citep{neuralMonkey}. The
parameters we used for the training are listed in
Appendix~\ref{sec:appendix:parameters}.  We will release the code upon
publication.

We used the AR baselines trained on the parallel data for generating
back-translated synthetic training data \citep{sennrich2016backtranslation}.
When training on back-translated data, authentic parallel data are upsampled to
match the size of the back-translated data.
We thus train our final models using the mix
of authentic and backtranslated data, so both AR baselines and the proposed
models use the same amount of data for training.
If we only used the parallel data
for training the neural models and kept the monolingual data only for the
language model, the proposed model would have benefited from having access to
more data than the AR baselines.

We train a 5-gram KenLM model \citep{heafield2011kenlm} on the monolignual data
tokenized using the same SentencePiece vocabulary as the parallel data.

For the perceptron training, we split the validation data for each language
pair in halves and use one half as the training set and the second half as a
held-out set. We use the score on the held-out set during the perceptron
training as an early-stopping criterion.  The scoring model is initialized with
zero weights for all features and a fixed weight of 1 for the CTC score.




\section{Results}%
\label{sec:results}

\begin{figure}
    \centering
    \input{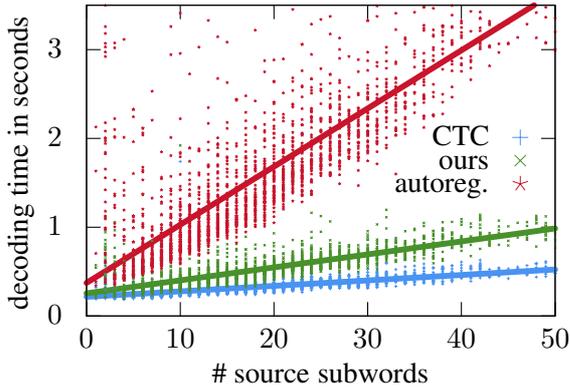}

    \caption{Comparison of the CPU decoding time of the autoregressive (AR),
    non-autoregressive (nAR) Transformer models and the proposed
    method with beam size of 10.}\label{fig:decoding_time}

\end{figure}

We evaluate our models on the standard WMT test sets that were previously used
for evaluation of nAR NMT\@. We use newstest2015 for English-German,
newstest2016 for English-Romanian, and newstest2018 for English-Czech
\citep{bojar2015findings, bojar2016findings, bojar2018findings}. We compute the
BLEU scores \citep{papineni2002bleu} as implemented in
SacreBLEU\footnote{https://github.com/mjpost/sacreBLEU} \citep{post2018call}.
We also measure the average decoding time for a single sentence.

Table~\ref{tab:results} shows the measured quantitative results of the
experiments.  We observe that the beam search greatly improves the translation
quality over the CTC-based nAR models (``Non-autoregressive'' vs. ``Ours'').
Additionally, we have control over the speed/quality trade-off by either
lowering or increasing the beam size.

Increasing the beam size from 1 to 5 systematically increases the translation
quality by at least 3 BLEU points. Decoding with a beam size of 20 matches the
quality of greedy autoregressive decoding while maintaining $1.5\times$
speedup.

Figure~\ref{fig:decoding_time} plots the time required to translate a sentence
with respect to its length.  As expected, beam search decoding is more
time-consuming than the CTC-based labeling (greedy). However, our method is
still substantially faster than the AR model, especially for longer
sentences.

Table~\ref{tab:ablstudy} shows how features used in the scoring model
contribute to the BLEU score.  We can see that combining the features is
beneficial and that the improvement is substantial with larger beam sizes. The
feature weights were trained separately for each beam size.

Our cursory manual evaluation indicates that additional features help to tackle
the most significant problems of nAR NMT -- repeated or malformed words and too
short sentences (see Appendix \ref{sec:appendix:examples} for examples).


\begin{table}

\begin{center}


\begin{adjustbox}{max width=0.95\columnwidth}
\begin{tabular}{lcccc}
\toprule
Beam Size &  1 &  5 &  10 &  20 \\ 
\midrule
$c+l+r+t$ & 22.68 & 25.50 & 25.93 & 26.03 \\
$c+l+r$   & 22.21 & 24.92 & 25.12 & 25.35 \\
$c+l$     & 22.05 & 24.64 & 24.77 & 25.12 \\
$c$       & 21.67 & 22.06 & 22.13 & 22.17 \\
\bottomrule
\end{tabular}
\end{adjustbox}

\end{center}

\caption{BLEU scores for English-to-German translation for different beam sizes
and feature sets: CTC score (\textit{c}), language model (\textit{l}), ratio
of the blank symbols (\textit{r}), and the number of trailing blank symbols
(\textit{t}).}
\label{tab:ablstudy}
\end{table}

\section{Related Work}%
\label{sec:related}


The earliest work on nAR translation includes work by
\citet{gu2017nonautoregressive} and \citet{lee2018deterministic} which are the
closest to our model beside our baseline. Unlike our approach, they do not
include state splitting.  \citet{gu2017nonautoregressive} used a latent
fertility model to copy a sequence of embeddings which is then
used for the target sentence generation.  \citet{lee2018deterministic} use two
decoders. The first decoder generates a candidate translation, which is then
iteratively refined by the second decoder a denoising auto-encoder or a masked
LM \citep{ghazvininejad-etal-2019-mask}.

\citet{junczys2018marian} exploit the autoregressive architectures
\citep{bahdanau2015neural,vaswani2017attention} and try to optimize the
decoding speed. Using model quantization and state memoization they achieve a
two-times speedup.






\section{Conclusions}%
\label{sec:conclusions}

We introduced a MT model with beam search that combines nAR CTC-based NMT model
with an $n$-gram LM and other features.

We performed experiments on six language pairs and evaluated the models on the
standard WMT sets.  Our approach narrows the quality gap between the nAR and AR
models while still maintaining a substantial speedup.

The experiments show that the main benefit of the proposed approach is the opportunity to balance the trade-off between translation quality and translation speed. The autoregressive models are still superior in translation quality for most of the language pairs, even though by a narrow margin. In contrast, the non-autoregressive models are very fast, but often lack in
translation quality. Our approach enhances constant-time neural network run with
a fast beam search utilizing a scoring model to improve the translation quality. By altering the beam size, we can adjust the speed and the quality ratio to achieve acceptable results both in terms of speed and translation quality.

\section*{Acknowledgements}

This research has been supported by the from the European Union's Horizon 2020 research and innovation programme under grant agreement No.~825303 (Bergamot), 
Czech Science Foundtion grant No.~19-26934X (NEUREM3), and Charles University grant No.~976518,
and has been using language resources distributed by the LINDAT/CLARIN project of the Ministry of Education, Youth and Sports of the Czech Republic (LM2015071). This research was partially supported by SVV project number 260~453.




\bibliography{main}
\bibliographystyle{acl_natbib}

\appendix

\newpage

\vspace*{10cm}

\newpage

\section{Appendix: Parameters}%
\label{sec:appendix:parameters}

The autoregressive baseline models use roughly the same set of hyperparameters
as the Transformer \emph{base} model \citep{vaswani2017attention}. Encoder and
decoder have 6 layers each, model dimension is 512, and the dimension of the
feed-forward layer is 2,048. We use 16 attention heads in both self-attention
and encoder-decoder attention. During training, we use label smoothing of 0.1
and we use dropout rate of 0.1. We use Adam optimizer \citep{kingma2014adam}
with parameters $\beta_1 = 0.9$, $\beta_2 = 0.997$, and $\epsilon = 10^{-9}$
with fixed learning rate of $10^{-4}$. Due to the GPU memory limitations, we
use batches of 20 sentences each, but we accumulate the gradients and perform
the only update the model parameters every  10 steps. (This makes our batch to
have an effective size of 200 sentences.) 

The hyperparameters of the CTC-based models were selected to be as comparative
as possible to the autoregressive models, with the following exceptions. The
splitting factor between the encoder and the decoder was selected to be $k =
3$, following the setup of \citet{libovicky2018end}. We lowered the number of
attention heads between the encoder and the decoder to 8 instead of 16. We changed the 
hyperparameter because it lead to better results in preliminary experiments.  
For training, instead of batching by a fixed number of sentences, we use batches of
maximum size of 400 tokens. We use the same delayed update interval of 10 steps
per update.

\newpage

\section{Appendix: Additional Results}\label{sec:appendix:results}

Quantitative results without the use back-translation, i.e., when the monolingual
data are used only for training the target-side language model are shown in in Table~\ref{tab:nobt}.

Quantitative results on WMT14 English-to-German Data for comparison with related work
are presented in Table~\ref{tab:wmt14}.

\begin{table}[!h]

    \centering
    \begin{tabular}{lcc}
    \toprule
    \multirow{2}{*}[-2pt]{Method} &
    \multicolumn{2}{c}{German WMT14} \\
    \\ \cmidrule(lr){2-3}

    & en $\rightarrow$ de
    & de $\rightarrow$ en

    \\ \midrule
    
    Non-autoregressive
      & 19.55   
      & 23.04   
    \\
    Transformer, greedy
      & 27.29   
      & 31.06 
    \\
    Transformer, beam 5
      & 27.71 
      & 31.85 
    \\
    \midrule

    Ours, beam 1
    & 20.59  
    & 24.11  

      \\
    Ours, beam 5
    & 23.61 
    & 27.19  
      \\
    Ours, beam 10
    & 24.27  
    & 27.83  

      \\
    Ours, beam 20
    & 24.41  
    & 28.14  

      \\
    \bottomrule
    \end{tabular}
    
    \caption{Quantitative results of the models in terms of BLEU on the WTM14 data.}
    \label{tab:wmt14}
    
\end{table}

\section{Appendix: Examples}
\label{sec:appendix:examples}

We include a few selected examples from the English-to-German (Table~\ref{tab:examples:ende}), German-to-English (Table~\ref{tab:examples:deen}), and Czech-to-English (Table~\ref{tab:examples:csen}) system outputs.

\begin{table*}[bp]
    \begin{tabular}{lccccccc}
    \toprule
    \multirow{2}{*}[-2pt]{Method} &
    \multicolumn{2}{c}{German WMT15} &
    \multicolumn{2}{c}{Romanian WMT16} &
    \multicolumn{2}{c}{Czech WMT18} &
    \multirow{2}{2cm}[-2pt]{\centering Decoding time [ms]}
    \\ \cmidrule(lr){2-3} \cmidrule(lr){4-5} \cmidrule(lr){6-7}

    & en $\rightarrow$ de
    & de $\rightarrow$ en
    & en $\rightarrow$ ro
    & ro $\rightarrow$ en
    & en $\rightarrow$ cs
    & cs $\rightarrow$ en

    \\ \midrule

      Non-autoregressive
      & 19.71   
      & 21.64   
      & 18.45   
      & 25.48   
      & 13.92   
      & 14.87   
      & \O314
    \\
    Transformer, greedy
      
      & 26.39   
      & 28.56   
      & 19.91   
      & 27.33   
      & 16.00   
      & 22.72   
      & 1637
      
    \\
    Transformer, beam 5
      & 26.99  
      & 29.39  
      & 20.81  
      & 27.99  
      & 17.08  
      & 23.54  
      & 4093
    \\
    \midrule

    Ours, beam 1
    & 20.81  
    & 22.68  
    & 18.45  
    & 26.52  
    & 14.86  
    & 16.11  
    & \O326

      \\
    Ours, beam 5
    & 23.29  
    & 25.96  
    & 20.88  
    & 29.67  
    & 17.16  
    & 20.87  
    & \O398
      \\
    Ours, beam 10
    & 23.99  
    & 26.19  
    & 21.52  
    & 29.88  
    & 17.20  
    & 21.52  
    & \O518
      \\
    Ours, beam 20
    & 24.01  
    & 26.59  
    & 22.02  
    & 29.94  
    & 17.24  
    & 21.87  
    & 1162

    \\ \bottomrule
    
\end{tabular}

\caption{Quantitative results in terms of BLEU \emph{without} the use of back-translation.}
\label{tab:nobt}

\end{table*}

\newcommand{\redund}[1]{{\color{black!30!red!100}\underline{#1}}}
\newcommand{\greenund}[1]{{\color{black!40!green!100}\underline{#1}}}
\newcommand{\blueund}[1]{{\color{black!30!blue!100}\underline{#1}}}

\begin{table*}[ht]

\begin{adjustbox}{max width=\textwidth}
\begin{tabular}{lp{16cm}}
    \toprule
   Source & On account of their innate aggressiveness, songs of that sort
were no longer played on the console.  \\

   nAR & Aufgrund \redund{ihrergeboren} \blueund{Aggressivitätivität} wurden Lieder dieser Art nicht mehr auf der Konsole gespielt. \\
   & \quad\rotatebox[origin=c]{180}{$\Lsh$} \it Two unrelated words are connected (red), malformed word with repeated subwords (blue). \\
   
   nAR + LM & Aufgrund ihrer \greenund{angeborenen} Aggressivität wurden Lieder dieser Art nicht mehr auf der Konsole gespielt \\
   & \quad\rotatebox[origin=c]{180}{$\Lsh$} \it Correct but too literal adjective was chosen. \\
   
   AR & Aufgrund ihrer \greenund{angeborenen} Aggressivität wurden Songs dieser Art nicht mehr auf der Konsole gespielt. \\
   
   Reference & Aufgrund ihrer ureigenen Aggressivität wurden Songs dieser Art nicht
mehr auf der Konsole gespielt. \\
   \midrule
   
   Source & Ailinn didn't understand. \\
   nAR &  \noindent \redund{A} hat nicht. \hspace{3.2cm} \it $\rightarrow$ Fail to copy infrequent proper name. \\
   
   nAR + LM & \noindent \redund{Aili} hat nicht verstanden.
   \hspace{1.1cm} $\rightarrow$ \it Non-LM features ensured more text is copied, but still incorrect. \\
   
   AR &  Ailinn verstand es nicht. \hspace{1.25cm} $\rightarrow$ \it  Correct. \\
   
   Reference &  Ailinn verstand das nicht. \\
   \midrule
   
   Source & Further trails are signposted, which lead up towards Hochrhön
and offer an extensive hike. \\
   nAR & Weitere Wege \redund{sindschilder}, die nach Hochrhön \blueund{und eine ausgedehnte Wanderung}. \\
   & \quad\rotatebox[origin=c]{180}{$\Lsh$} \it Two unrelated words are connected (red), missing verb in the second clause (blue). \\

   nAR + LM & Weitere Wege sind ausgeschilder, die in Hochrhön \blueund{und eine ausgedehnte Wanderung}. \\
   & \quad\rotatebox[origin=c]{180}{$\Lsh$} \it Connected words got corrected, the second clause (blue) still does not make sense. \\
   
   AR &  Weitere Wege sind ausgeschildert, die in Richtung Hochrhön führen und eine ausgedehnte Wanderung bieten. \hspace{4cm} $\rightarrow$ \it Correct. \\
   
   Reference & Weitere Wege sind ausgeschildert, die Richtung Hochrhön hinaufsteigen
und zu einer ausgedehnten Wanderung einladen. \\
   
   \bottomrule
\end{tabular}
\end{adjustbox}

\caption{Manually selected examples of system outputs for English-to-German translation containing the most frequent error types. `nAR' is the purely non-autoregressive
    system, `nAR + LM' is the proposed system with beam size 20.}\label{tab:examples:ende}

\end{table*}

\begin{table*}[h!]
\begin{adjustbox}{max width=\textwidth}
\begin{tabular}{lp{16cm}}
\toprule
Source        & Aber diese Selbstzufriedenheit ist unangebracht. \\

nAR   & But \redund{comp} complacency is misguided. \\

nAR + LM & But complacency is misguided. \\

AR    & But this complacency is inappropriate. \\

Reference        & But such complacency is misplaced. \\

\midrule
Source        & Als ich also sehr, sehr übergewichtig wurde und Symptome von Diabetes zeigte, sagte mein Arzt "Sie müssen radikal sein.\\

nAR   & So when I \greenund{very, very} overweight \redund{and and} showed symptoms of diabetes, \redund{my my} doctor said "You must be radical.\\

nAR + LM & So when I became \greenund{very, very} overweight \greenund{and} showed symptoms of diabetes, \greenund{my} doctor said "You must be radical.\\

AR    & So when I was very, very overweight and showed symptoms of diabetes, my doctor said "You must be radical.\\

Reference        & So when I became very, very overweight and started getting diabetic symptoms, my doctor said, 'You've got to be radical.\\

\bottomrule
\end{tabular}
\end{adjustbox}

\caption{German-to-English examples.}
\label{tab:examples:deen}
\end{table*}

\begin{table*}[h!]
\begin{adjustbox}{max width=\textwidth}
\begin{tabular}{lp{16cm}}
\toprule
Source        & Problémem mohou být také jednorázové pleny. \\

nAR   & \noindent{\color{black!30!red!100}\underline{Singleaperslso}} be problem. \\

nAR + LM & \noindent{\color{black!30!red!100}\underline{One can}} diapers be the problem. \\

AR    & Single diapers may also be the problem.\\

Reference        & Disposable incontinence pants may also be a problem. \\

\midrule

Source        & Pere se ve mně adolescentní potřeba uchechtnout se s obdivem nad tím, s jakým vážným tónem je mi výklad podáván. \\

nAR   & I adolescent need {\color{black!30!red!100}\underline{tohuck}} with admiration the serious tone my interpret. \\

nAR + LM & I {\color{black!40!green!100}\underline{have}} a adolescent need to chuck with {\color{black!30!red!100}\underline{wonderation}} of the serious tone my interpret. \\

AR    & I\redund{'m asking for} an adolescent need to laugh at the admiration of the serious tone of my interpretation.\\

Reference        & I feel the adolescent need to chuckle with admiration for the serious tone with which my comment is handled. \\
\bottomrule

\end{tabular}
\end{adjustbox}
\caption{Czech-to-English examples.}
\label{tab:examples:csen}
\end{table*}

\end{document}